\documentclass{article} 
\usepackage[preprint]{neurips_2021}
\usepackage[utf8]{inputenc} 
\usepackage[T1]{fontenc}    
\usepackage{booktabs}       
\usepackage{amsfonts}       
\usepackage{nicefrac}       
\usepackage{microtype}      
\usepackage{xcolor}         
\usepackage{times}
\usepackage[colorlinks = true,
            linkcolor = blue,
            urlcolor  = blue,
            citecolor = blue,
            anchorcolor = blue]{hyperref}      
\usepackage{url}
\usepackage{latexsym}
\usepackage{amsmath}
\usepackage{graphicx} 
\usepackage{multirow}
\usepackage{multicol}
\usepackage{algorithm}    
\usepackage{algpseudocode}
\usepackage{amssymb}
\usepackage{ucs}
\usepackage{enumitem}
\usepackage{color}
\usepackage{subcaption}
\usepackage{amsmath}
\usepackage{amstext}
\usepackage{amsfonts}
\usepackage{bm}
\usepackage{hyperref}
\usepackage{geometry}
\usepackage{pgfplots}
\usepackage{fancyhdr}
\usepackage{soul}

\usepackage{bbm}
\usepackage{multirow}
\title{Inference with Reference: \\ Lossless Acceleration of Large Language Models}

\author{Nan Yang, ~~Tao Ge, ~~Liang Wang, ~~Binxing Jiao \\
 \textbf{Daxin Jiang,} ~~\textbf{Linjun Yang,} ~~\textbf{Rangan Majumder,} ~~\textbf{Furu Wei}   \\
 Microsoft\\
\texttt{\{nanya, tage, wangliang, binxjia,} \\
\texttt{djiang, linjya, ranganm, fuwei\}@microsoft.com} \\
\url{https://github.com/microsoft/unilm}
}

\begin{document}

\maketitle

\begin{abstract}
We propose \textbf{LLMA}, an LLM accelerator to losslessly speed up Large Language Model (LLM) inference with references. LLMA is motivated by the observation that there are abundant identical text spans between the decoding result by an LLM and the reference that is available in many real world  scenarios (e.g., retrieved documents). LLMA first selects a text span from the reference and copies its tokens to the decoder and then efficiently checks the tokens' appropriateness as the decoding result in parallel within one decoding step.
The improved computational parallelism allows LLMA to achieve over $2\times$ speed-up for LLMs with identical generation results as greedy decoding in many practical generation scenarios where significant overlap between in-context reference and outputs exists (e.g., search engines and multi-turn conversations).
\end{abstract}
\begin{figure}[htbp]
    \centering
    \begin{subfigure}{0.3\textwidth}
     \centering
     \includegraphics[width=1.0\textwidth]{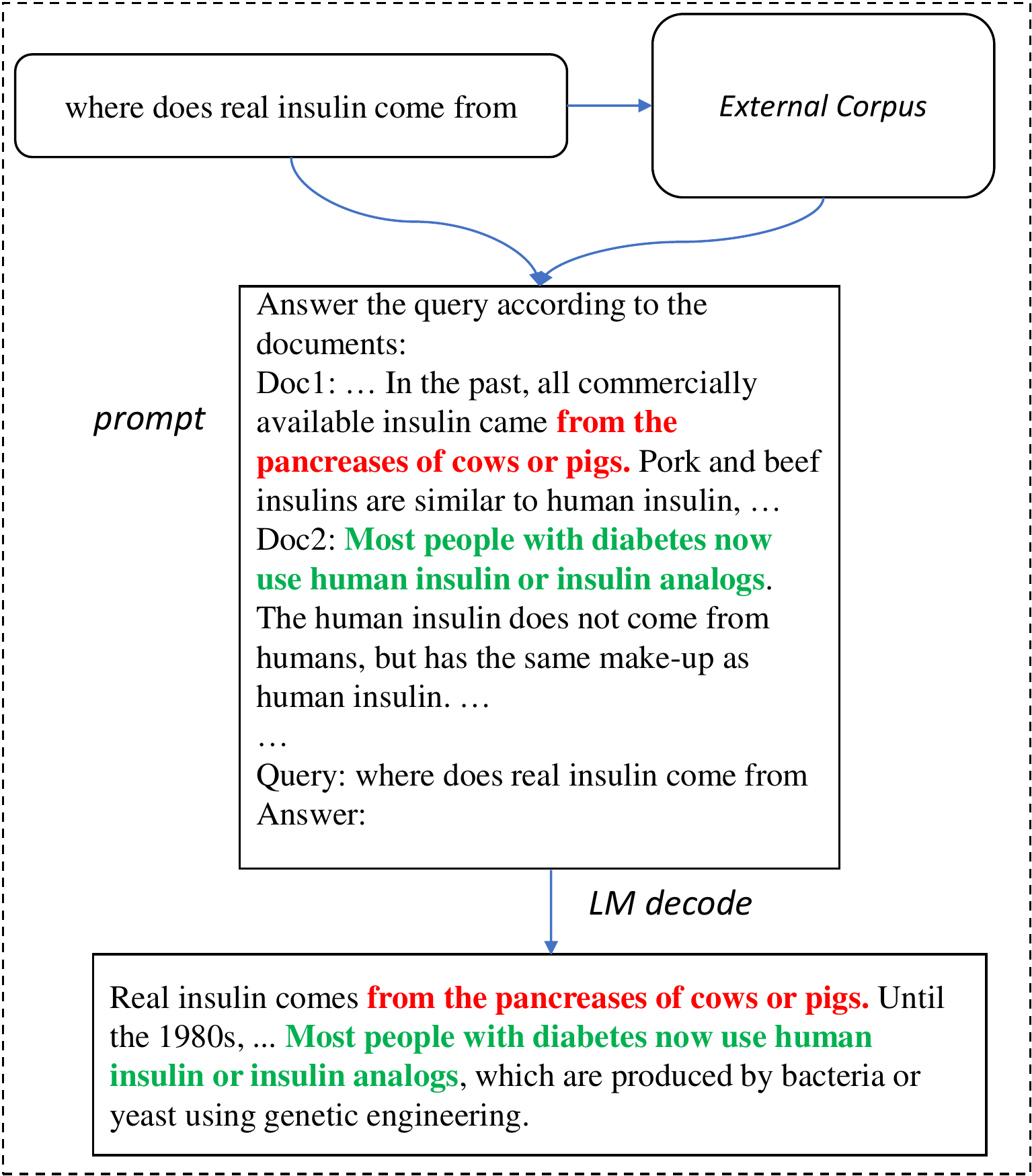}
     \caption{Retrieval-augmented.}
     \label{subfig:rag}
    \end{subfigure}
    \begin{subfigure}{0.3\textwidth}
     \centering
     \includegraphics[width=1.0\textwidth]{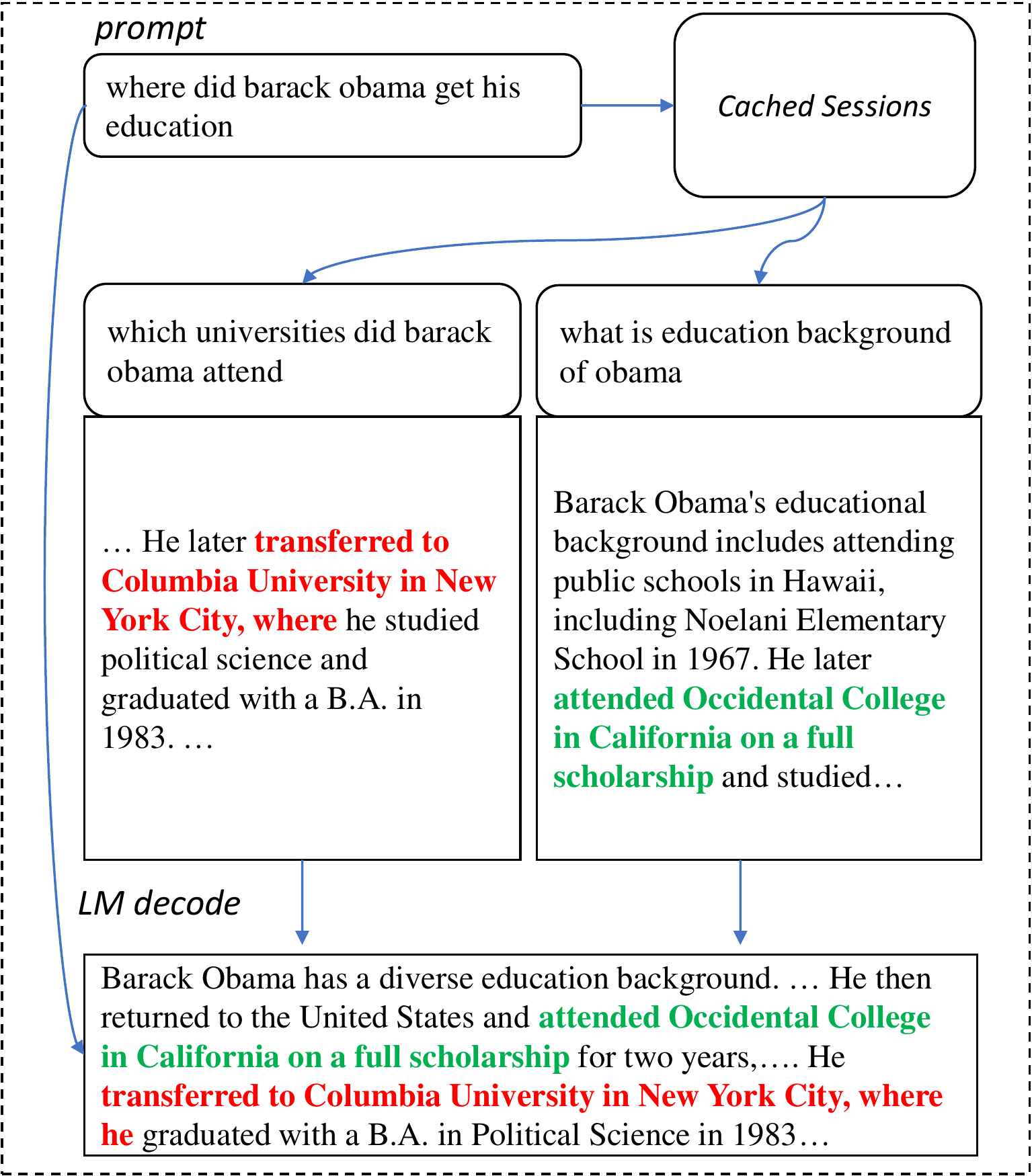}
     \caption{Cache-assisted.}
     \label{subfig:cag}
    \end{subfigure}
    \begin{subfigure}{0.3\textwidth}
     \centering
     \includegraphics[width=1.0\textwidth]{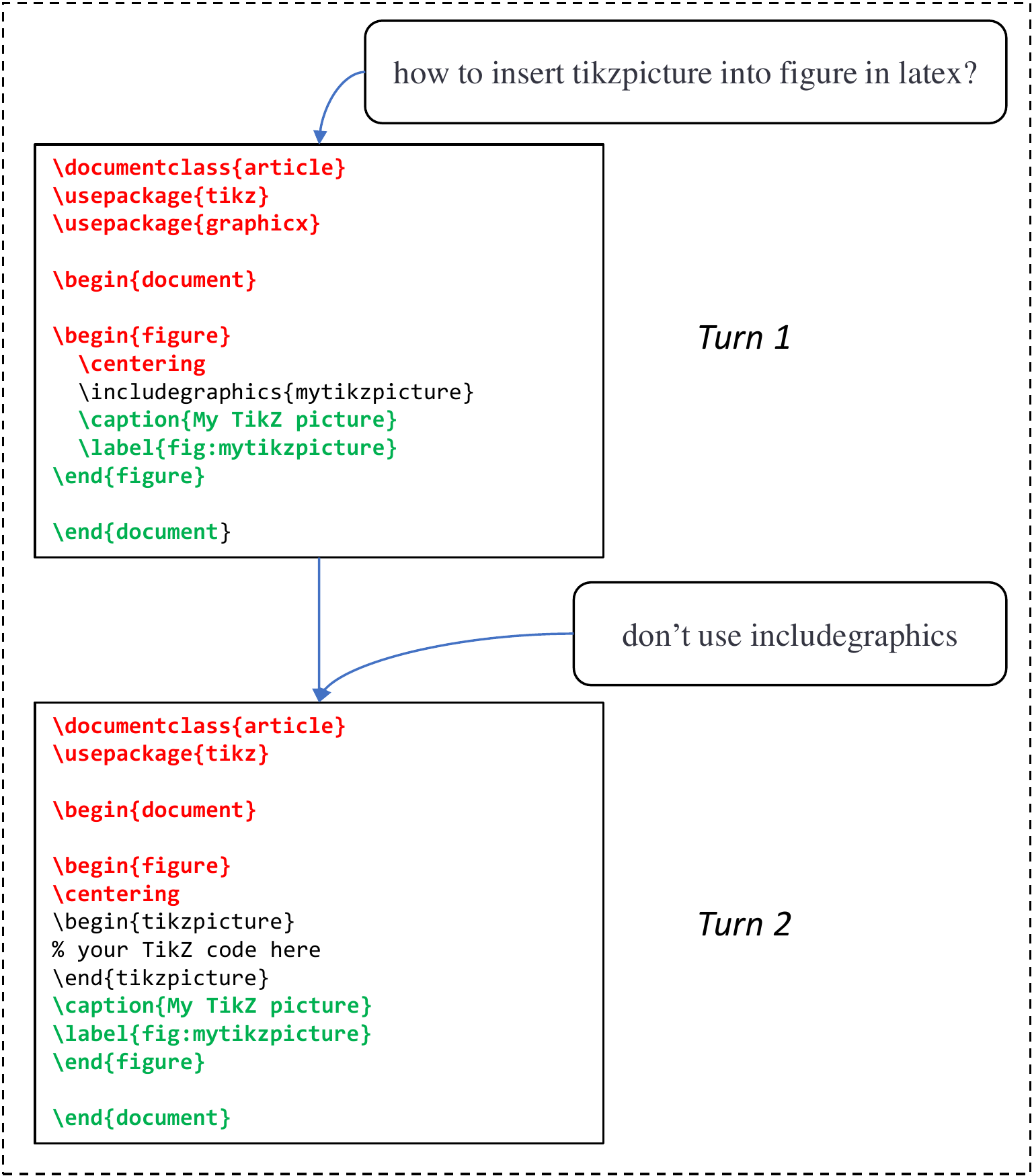}
     \caption{Multi-turn conversations.}
     \label{subfig:mtt}
    \end{subfigure}
    
    \caption{Significant overlaps between inputs and references exist in many LLM applications such as retrieval-augmented generation, cache-assisted generation and multi-turn conversations. By exploiting such overlaps, our \textbf{LLMA} method can accelerate the inference of LLMs up-to 2$\sim$3 times without additional models.}
    \label{fig:task-overall}
\end{figure}

\section{Introduction}\label{sec:intro}
With large foundation models (e.g., GPT-3.5/GPT-4) \citep{openai2023gpt4} becoming widely used for various real-world applications, the concern of high deployment cost has been increasingly raised. While there are general methodologies that help reduce the serving cost of LLMs such as quantization \citep{dettmers2023case}, pruning \citep{frantar2023sparsegpt}, compression \citep{xu2020bert} and distillation \citep{wang2020minilm}, the inference efficiency bottleneck of these transformer-based generative models (e.g., GPT) is mainly associated with autoregressive decoding: at test time, output tokens must be decoded (sequentially) one by one, which poses
significant challenges for the LLMs to be deployed at scale.

In this work, we study accelerating LLM's inference by improving the efficiency of autoregressive decoding. In many real world applications, we observe that an LLM's output tokens often come from its context. For example, in a typical retrieval-augmented generation scenario for a search engine, an LLM's context usually includes relevant documents that are retrieved from an external corpus as reference according to a query, and its output usually contains many text spans found in the reference (i.e., retrieved documents), as shown in Figure \ref{fig:task-overall}.

Motivated by the above observation, we propose \textbf{LLMA}, an inference-with-reference decoding mechanism to accelerate LLM inference by exploiting the overlap between an LLM's output and reference that is available for many practical scenarios. LLMA first selects a text span from the reference and copies its tokens to the LLM decoder and then checks if they are acceptable based on the output token probabilities, which can be conducted efficiently in parallel.
In this way, we can accelerate decoding by enabling better parallelism on vector accelerators such as GPUs while ensuring the generation results are identical to the vanilla greedy decoding method.

Compared to previous efficient decoding algorithms such as Speculative Decoding\footnote{It was named Generalized Aggressive Decoding in the early version of the manuscript~\citep{xia2022lossless}.} \citep{xia22speculative} and Speculative Sampling \citep{chen2023accelerating} that need to introduce an additional efficient drafter model to generate a draft for checking, LLMA does not require an additional model and is easy to implement and deploy, which is an extension of our previous work -- (Input-guided) Aggressive Decoding \citep{sun2021instantaneous,ge2022lossless} that demonstrates a success in the rewriting tasks (e.g., Grammatical Error Correction) where inputs and outputs are similar. 

Experiments show that our LLMA method can generate identical results as greedy decoding but achieve over $2\times$ speed-up across different model sizes in practical application scenarios like retrieval-augmented and cache-assisted generation.

\begin{figure}[htbp]
    \centering
    \includegraphics[width=1.0\textwidth]{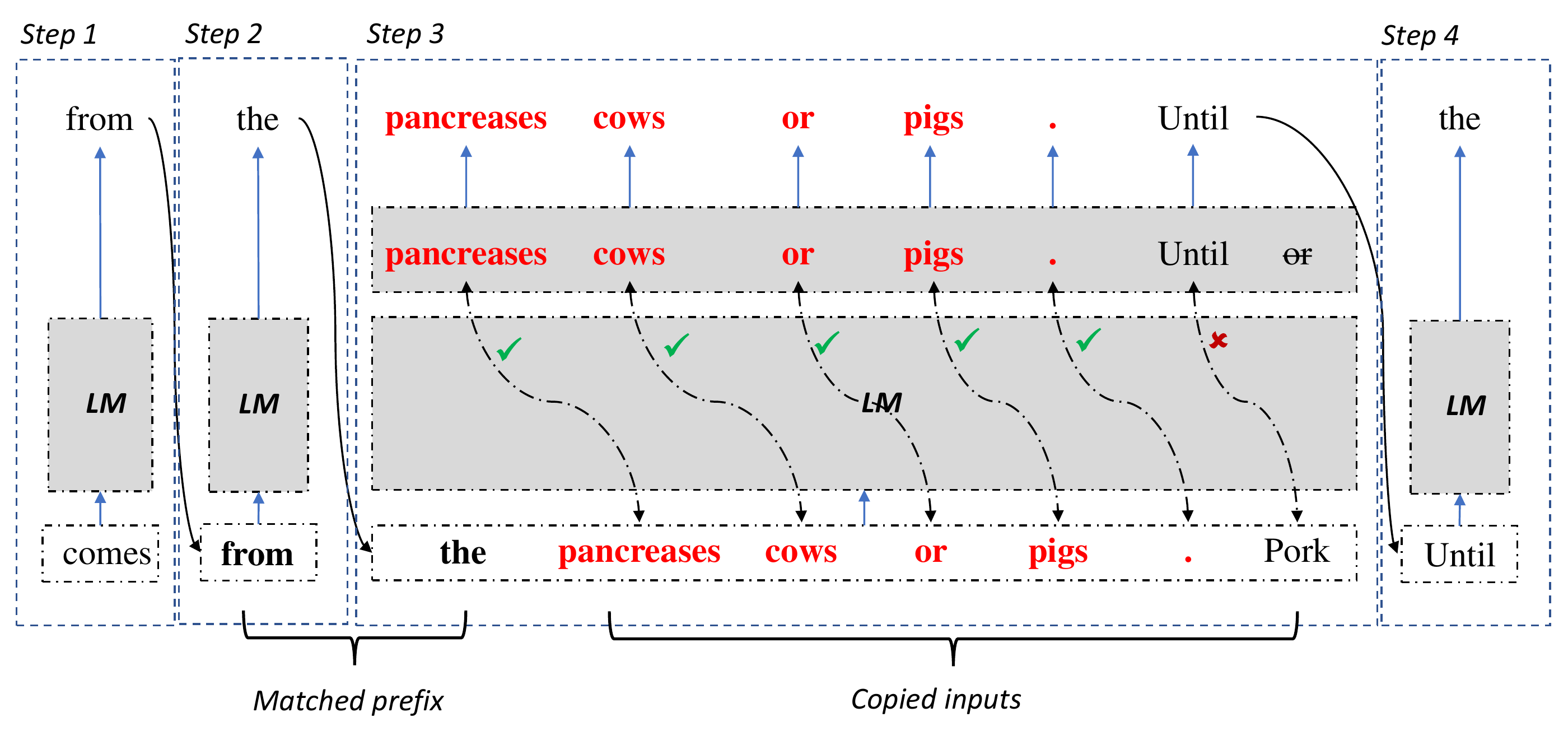}
    \caption{Illustration of \textbf{LLMA} decoding algorithm. At step 3, copy mechanism is triggered as ``\textit{from the}'' is matched against some reference document (see Figure \ref{fig:task-overall}a for details). The text span ``\textit{pancreases cows or pigs . Pork}'' is copied from the reference document into the input of the LM decoder. The copied tokens are then efficiently checked by running the LM to compute their output tokens in one decoding step. ``\textit{pancreases cows or pigs .}'' are identical to their previous decoding output tokens and accepted as valid inputs, while the last input token ``\textit{Pork}'' does not match the previous output token ``\textit{Until}'' and is thus invalid and discarded. Overall, at step 3, \textbf{LLMA} generates six new output tokens ``\textit{pancreases cows or pigs . Until}'' compared to one output token per step for the baseline decoding algorithm.}
    \label{fig:decode-tetail}
\end{figure}

\section{Method}\label{sec:method}
\subsection{Background: Stepwise Decoding in Autoregresive Language Models}
Autoregressive language models typically follow a step-by-step decoding algorithm. Let $\boldsymbol{x}$ be the user given prompt sequence, $\boldsymbol{y}$ be the LM output sequence.
At each decoding step $i$, the model takes the concatenation of $\boldsymbol{x}$ and previous generated tokens $\boldsymbol{y}_{<i}$ as input and computes the probability of the next token over the LM vocabulary $V$:
\begin{equation}
    y_{i} \sim P(y|\boldsymbol{x},\boldsymbol{y}_{< i})
\end{equation}
For greedy-decoding, we always select the token with the highest probability as the output:
\begin{equation}\label{eq:stepwise-decoding}
    y_{i} = \arg \max_{y\in V} P(y|\boldsymbol{x},\boldsymbol{y}_{< i})
\end{equation}
$y_{i}$ is then appended to $\boldsymbol{y}_{<i}$ and the process is repeated until the full sequence is generated.
Each decoding step depends on the results of previous steps and the dependency slows down the decoding process as it hampers the full utilization of parallel computing on vector accelerators such as GPUs.

\subsection{LLMA: Inference with Reference}
Let $\boldsymbol{x}$ be the input prompt, $\boldsymbol{y}$ be the generation output and $\boldsymbol{D}$ be a list of reference documents.
We assume the generation result $\boldsymbol{y}$ and the reference documents $\boldsymbol{D}$ have many identical text spans. 
We exploit this redundancy to accelerate the decoding of language models by copying texts from $\boldsymbol{D}$ into the input of LM decoder.
Similar to the idea of suffix matching in our previous work \citep{sun2021instantaneous}, we check if the previously generated $n$ tokens $\boldsymbol{y}_{i-n:i}$ match some text spans in $\boldsymbol{D}$ at each decoding step $i$.
If no match is found, we fall back to stepwise decoding described in Equation~\ref{eq:stepwise-decoding}. If multiple matching spans are found, we select the span with the longest matching prefixes with $\boldsymbol{y}_{<i}$. We randomly break the tie if multiple spans exist with the longest matching prefixes. If span $(pos-n,pos)$ in reference document $\boldsymbol{d}$ is matched, we assume the model output is likely to follow $\boldsymbol{d}$. Thus we preemptively copy $k$ subsequent tokens $\boldsymbol{d}_{pos:pos+k}$ from $\boldsymbol{d}$ to the language model input. Given $\boldsymbol{y}_{<i}$ and $\boldsymbol{d}_{pos:pos+k}$, the model can produce $k+1$ output tokens in a single decoding step:
\begin{equation}
    y_{i+j} = \arg \max_{y\in V} P(y|\boldsymbol{x},\boldsymbol{y}_{<i},\boldsymbol{d}_{pos:pos+j}), j=0,..,k
\end{equation}
For $y_{i+j}$ to be valid output, we must ensure all tokens in its input are valid, which means copied tokens $\{d_{pos},...d_{pos+j-1}\}$ are identical to previous model output $\{y_i,...,y_{i+j-1}\}$.
The invalid output tokens are discarded and the valid output tokens are appended to the output sequence $\boldsymbol{y}$.
The verification guarantees our decoding algorithm produces identical generation output as the stepwise decoding method for greedy decoding.
Furthermore, LLMA can generate between $1$ and $k+1$ output tokens per decoding step, compared to only one token per step for the stepwise decoding method. See Algorithm~\ref{algo:llma} for the pseudo code and Figure~\ref{fig:decode-tetail} for the illustration of our method.

\begin{algorithm}[t]
\caption{LLMA Decoding.}\label{algo:llma}
\textbf{Input:} $\boldsymbol{x}$, $\boldsymbol{D}=(\boldsymbol{d}_1,\dots,\boldsymbol{d}_n)$, $n$, $k$, $N$; \\
\textbf{Output:} $\boldsymbol{y}$;
\begin{algorithmic}[1]
\State $\boldsymbol{y} \gets []$
\While{$\Call{len}{\boldsymbol{y}} < N$}
\State $matched, \boldsymbol{d}, pos \gets \Call{match\_ngrams}{\boldsymbol{y}, \boldsymbol{D}, n}$

\If{$\neg matched$}
\State $o = \Call{LLM}{\boldsymbol{x},\boldsymbol{y}}$ 
\State $\Call{append}{\boldsymbol{y}, o}$
\State \textbf{continue}
\EndIf
\State $(o_0,o_1,\dots,o_k) \gets \Call{LLM}{\boldsymbol{x}, \boldsymbol{y}, d_{pos},...,d_{pos+k-1}}$
\State $\Call{append}{\boldsymbol{y}, o_0}$
\For{$i \textbf{ in } 0,\dots,k-1$}
\If{$o_{i} \textbf{ neq } d_{pos+i} $}
\State \textbf{break}
\EndIf
\State $\Call{append}{\boldsymbol{y},o_{i+1}}$
\EndFor
\EndWhile
\end{algorithmic}
\end{algorithm}

Overall, our decoding algorithm has two hyper-parameters: the match length $n$ and the copy length $k$, which control how aggressively we trigger and apply the copy mechanism.

\subsection{Application Scenarios}
Our decoding algorithm can be beneficially applied to any scenarios where the generation outputs have significant overlaps with reference documents. In this work, we discuss three important practical scenarios that LLMA can benefit: 

\noindent \textbf{Retrieval-Augmented Generation}. 
We let $\boldsymbol{q}$ be the user given query tokens.
In retrieval-augmented generation, a list of reference documents $\boldsymbol{D}$ are retrieved from an external corpus based on their relevance to the query $\boldsymbol{q}$. The documents $\boldsymbol{D}$ and the user query $\boldsymbol{q}$ are then inserted into some pre-defined template to create the final input $\boldsymbol{x}$ for the language model:
\begin{equation}
    \boldsymbol{x} = \texttt{Template}~(\boldsymbol{q},\boldsymbol{D})
\end{equation}
From $\boldsymbol{x}$, the language model generates output sequence $\boldsymbol{y}$, which usually contains significant overlaps with the reference documents $\boldsymbol{D}$. See Figure~\ref{subfig:rag} for an example.

\noindent \textbf{LLM Generation with Cached Sessions}.
When serving LLMs, previously generated sessions can be stored in a cache to speed up future generations.
The cache is a collection of past query-output pairs $(\boldsymbol{q},\boldsymbol{d})$. Given a new query $\boldsymbol{q}$, similar queries can be retrieved from the cache and their corresponding outputs can be used as references $\boldsymbol{D}$ in our LLMA method to accelerate decoding. Figure~\ref{subfig:cag} gives an example of cached-assisted generation.

\noindent \textbf{Multi-turn Conversation with LLMs}. New interaction patterns with LLMs emerge with the powerful LLMs such as GPT-4. One important pattern is that LLMs are repeatedly asked to refine their own outputs either by the users or the LLM themselves \citep{gao2022rarr}. Under such scenarios, the generation outputs between different turns often have significant overlaps, which can be exploited by treating previous turns as reference documents. See Figure~\ref{subfig:mtt} for an example.   

\section{Experiment}\label{sec:exp}
\subsection{Dataset}
Our dataset consists of triples of input prompt $\boldsymbol{x}$, target generation result $\boldsymbol{y}$ and reference documents $\boldsymbol{D}$.
Obtaining realistic generation outputs is crucial for meaningful evaluation of our method as the efficiency of our method depends on the overlaps between reference documents and the generation output.
Weak LLMs tend to produce either short, irrelevant output with little overlaps with references or repeat the reference documents with no modification, either underestimating or overestimating the efficiency of the proposed method.
While any strong LLMs would suffice, in this work, we opt to use the davinci-003 variant of GPT-3.5 model through OpenAI APIs \footnote{\url{https://platform.openai.com/docs/models/gpt-3-5}} to produce high-quality generation results.

\noindent \textbf{Retrieval-Augmented Generation (RAG)}. We start by sampling queries from the MS-MARCO passage retrieval dataset \citep{bajaj2018ms}.
For each query $\boldsymbol{q}$, we use a dual-encoder retrieval model E5 \citep{wang2022text} to retrieve a list of 10 passages $\{\boldsymbol{d}_i\}_{i=1}^{10}$ from the MS-MARCO corpus. Davinci-003 is prompted to generate a response $\boldsymbol{y}$ for the query according to the retrieved 10 passages. We then combine $\boldsymbol{q}$ and $\{\boldsymbol{d}_i\}_{i=1}^{10}$ to get the prompt $\boldsymbol{x}$ for our retrieval-augmented generation. See Appendix~\ref{ap_sec:prompt} for the detailed prompt templates.

\noindent \textbf{Cache-Assisted Generation (CAG)}. We reuse queries from MS-MARCO passage retrieval dataset. For each query $\boldsymbol{q}$, we use davinci-003 to generate 4 similar queries to simulate the cached queries. For both the original and the similar queries, davinci-003 is prompted to respond without reference to the retrieved documents. We treat the original query $\boldsymbol{q}$ as the input prompt $\boldsymbol{x}$, the response to the original queries as generation result $\boldsymbol{y}$, and the responses to the similar queries as the reference documents $\boldsymbol{D}$.

For both scenarios, we produce 200 triples of $(\boldsymbol{x}, \boldsymbol{y}, \boldsymbol{D})$. 100 triples are used as dev set for tuning hyper-parameters (match length $n$ and copy length $k$) and conducting ablation studies, and the other 100 triples are used for the final test evaluation. Table~\ref{tab:data_stats} shows the input and output lengths for samples in all datasets. The input prompts are significant longer in the RAG settings because the retrieved documents are inserted into the prompts for RAG while the cached sessions are not in the inputs for cached-assisted generation.
\begin{table}[ht]
  \centering
  \begin{tabular}{c|c|c|c|c}
 \hline
    \multirow{2}{*}{\textbf{\#tokens}} & \multicolumn{2}{c|}{\textbf{Retrieval}} & \multicolumn{2}{c}{\textbf{Cache}} \\
    \cline{2-5}
    & \textbf{dev} & \textbf{test} & \textbf{dev} & \textbf{test}  \\
    \hline
    Input & 903.6 & 898.8 & 15.6 &  17.1 \\
    Output & 111.2 & 122.0 & 162.5 & 177.3  \\
    \hline
  \end{tabular}
  \vspace{0.2cm}
  \caption{Numbers of input and output tokens per sample.}
  \label{tab:data_stats}
\end{table}

\subsection{Target Guided Simulation}
We test the proposed method using open sourced LLaMA \citep{touvron2023llama} language models.
Unfortunately, the outputs of LLaMA do not follow the generation results from davinci-003 model.
Fortunately, for greedy-decoding, the decoding process of our method can be fully inferred from the davinci-003 output $\boldsymbol{y}$ and the reference documents $\boldsymbol{D}$.
Given match length $n$ and copy length $k$, for every decoding step, we can determine the numbers of input and output tokens using Algorithm~\ref{alg:infer_ds}.
We can force LLaMA model to follow the exact decoding steps regardless of its own output, which is sufficient for measuring the execution time of our method.
\begin{algorithm}[t]
\caption{Infer decoding sequence from target sequence and reference documents.\label{alg:infer_ds}}
\textbf{Input:} $\boldsymbol{y}, \boldsymbol{D}=(\boldsymbol{d}_1,\dots,\boldsymbol{d}_n), n, k$; \\
\textbf{Output:} $\boldsymbol{s}=(i_1,o_1),\dots,(i_m,o_m)$;
\begin{algorithmic}[1]
\State $step \gets 0$
\State $\boldsymbol{s} \gets []$
\While{$step < \Call{len}{\boldsymbol{y}}$}
\State $matched, d, pos \gets \Call{match\_ngrams}{\boldsymbol{y}, step, D, n}$

\If{$\neg matched$}
\State $step \gets step + 1$
\State $\Call{append}{\boldsymbol{s}, (1, 1)}$
\State \textbf{continue}
\EndIf
\State $num\_valid \gets \Call{get\_matched\_tokens}{d, pos, \boldsymbol{y}, step}$

\State $num\_valid \gets \Call{min}{k, num\_valid}$
\State $num\_output\_tokens \gets num\_valid + 1$
\State $step \gets step + num\_output\_tokens$ 
\State $\Call{append}{\boldsymbol{s}, (1+k, num\_output\_tokens)}$
\EndWhile
\end{algorithmic}
\end{algorithm}

\subsection{Implementation Details}
We use the Huggingface Transformers library \citep{wolf-etal-2020-transformers} to implement the inference for both the autoregressive decoding baseline and our LLMA decoding method. We use the \textit{accelerate} library \citep{accelerate} to implement larger models sharded to multiple GPUs. We perform tests on LLaMA model of 7B, 13B and 30B parameters. All the inferences are done in half floating numbers. For the 7B and 13B models, the inferences are done in one NVidia 32G V100 GPU, and for the 30B model, the inference is performed on four NVidia 32G V100 GPUs on a single machine. All inferences are done with greedy-decoding, with batch size 1. 

\subsection{Main Results}
We determine the match length $n$ and copy length $k$ by running grid search on the dev set. Table~\ref{tab:kn_stat} shows the optimal $n$ and $k$ values for different scenarios and different model sizes. We then run our experiments on the test set for three rounds and the averaged results are shown in Table~\ref{tab:main_results_retrieval} and Table~\ref{tab:main_results_cache}. Our LLMA method achieves 2 to 3 times speed-up over baseline across different model sizes and scenarios.

\begin{table}[h]
  \centering
  \begin{tabular}{c|c|c|c|c}
 \hline
    \multirow{2}{*}{\textbf{Model}} & \multicolumn{2}{c|}{\textbf{Retrieval}} & \multicolumn{2}{c}{\textbf{Cache}} \\
    \cline{2-5}
    & $\boldsymbol{n}$ & $\boldsymbol{k}$ & $\boldsymbol{n}$ & $\boldsymbol{k}$  \\
    \hline
    7B & 1 & 18 & 1 &  15 \\
    13B & 1 & 14 & 1 & 15  \\
    30B & 1 & 18 & 1 & 18  \\
    \hline
  \end{tabular}
  \vspace{0.2cm}
  \caption{Match length $n$ and copy length $k$ determined by grid search on dev set.}
  \label{tab:kn_stat}
\end{table}

\begin{table}[h]
  \centering
  \begin{tabular}{c|c|c|c|c|c}
 \hline
    \multirow{2}{*}{\textbf{Model}} & \multicolumn{2}{c|}{\textbf{Tokens/sec} $\uparrow$} & \multicolumn{2}{c|}{\textbf{Time (sec)} $\downarrow$} & {\multirow{2}{*}{\textbf{Speed-up} $\uparrow$}} \\
    \cline{2-5}
    & \textbf{baseline} & \textbf{LLMA} & \textbf{baseline} & \textbf{LLMA} & \\
    \hline
    7B & 23.9 & 59.2 & 511.2 & 206.0 & 2.48x \\
    13B & 18.5 & 41.1 & 658.4 & 296.8 & 2.22x \\
    30B & 4.9 & 12.1 & 2503.2& 1005.8 & 2.49x \\
    \hline
  \end{tabular}
  \vspace{0.2cm}
  \caption{Time comparison for retrieval-augmented generation. The times are the total execution times in seconds of decoding 100 samples. All numbers are averaged over 3 runs.}
  \label{tab:main_results_retrieval}
\end{table}

\begin{table}[h]
  \centering
  \begin{tabular}{c|c|c|c|c|c}
 \hline
    \multirow{2}{*}{\textbf{Model}} & \multicolumn{2}{c|}{\textbf{Tokens/sec} $\uparrow$} & \multicolumn{2}{c|}{\textbf{Time (sec)} $\downarrow$} & {\multirow{2}{*}{\textbf{Speed-up $\uparrow$}}} \\
    \cline{2-5}
    & \textbf{baseline} & \textbf{LLMA} & \textbf{baseline} & \textbf{LLMA} & \\
    \hline

    7B & 24.3 & 53.8 & 730.8 & 329.8 & 2.22x \\
    13B & 19.3 & 42.3 & 918.4 & 419.3 & 2.19x \\
    30B & 5.1 & 15.6 & 3467.7 & 1133.0 & 3.06x \\
    \hline
  \end{tabular}
  \vspace{0.2cm}
  \caption{Time comparison for generation with cached sessions. The times are the total execution times in seconds of decoding 100 samples. All numbers are averaged over 3 runs.}
  \label{tab:main_results_cache}
\end{table}

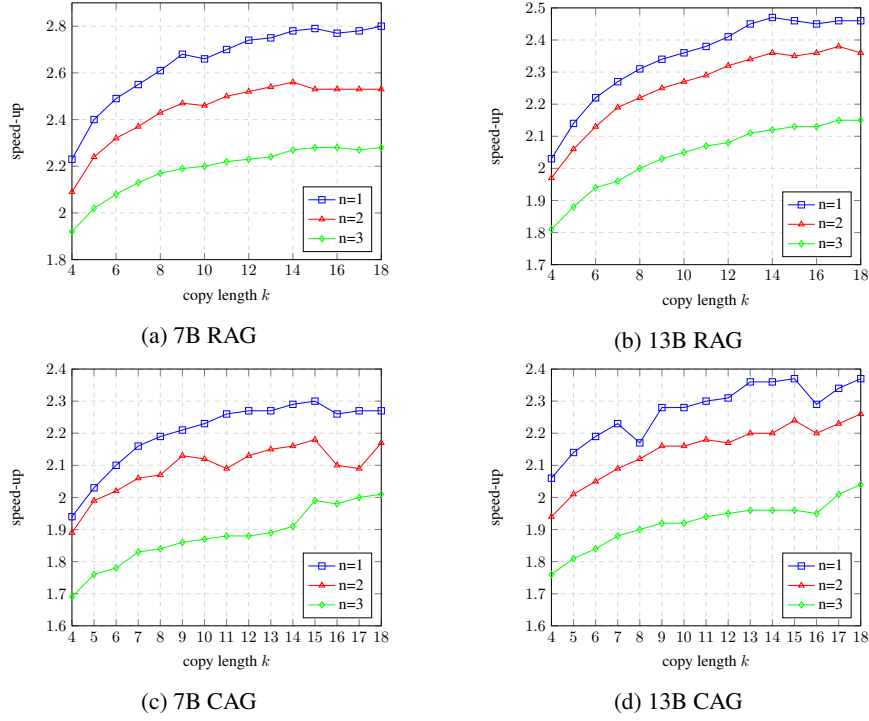
\begin{figure}[htbp]
    \centering
    \begin{subfigure}{0.45\textwidth}
        \centering
        \begin{tikzpicture}[scale=0.6]
            \begin{axis}[
    xlabel={copy length $k$},
    ylabel={speed-up},
    xmin=4, xmax=18,
    ymin=1.8, ymax=2.9,
    xtick={4,6,...,18},
    ytick={1.8,2.0,...,2.8},
    grid=major,
    grid style={dashed,gray!30},
    legend pos=south east,
]

\addplot[
    color=blue,
    mark=square,
    ]
    coordinates {
    (4,2.23)(5,2.4)(6,2.49)(7,2.55)(8,2.61)(9,2.68)(10,2.66)(11,2.7)(12,2.74)(13,2.75)(14,2.78)(15,2.79)(16,2.77)(17,2.78)(18,2.8)
    };
    \addlegendentry{n=1}

\addplot[
    color=red,
    mark=triangle,
    ]
    coordinates {
    (4,2.09)(5,2.24)(6,2.32)(7,2.37)(8,2.43)(9,2.47)(10,2.46)(11,2.5)(12,2.52)(13,2.54)(14,2.56)(15,2.53)(16,2.53)(17,2.53)(18,2.53)
    };
    \addlegendentry{n=2}

\addplot[
    color=green,
    mark=diamond,
    ]
    coordinates {
    (4,1.92)(5,2.02)(6,2.08)(7,2.13)(8,2.17)(9,2.19)(10,2.2)(11,2.22)(12,2.23)(13,2.24)(14,2.27)(15,2.28)(16,2.28)(17,2.27)(18,2.28)
    };
    \addlegendentry{n=3}

\end{axis}
        \end{tikzpicture}
        \caption{7B RAG}
    \end{subfigure}
    \centering
    \begin{subfigure}{0.45\textwidth}
        \centering
        \begin{tikzpicture}[scale=0.6]
            \begin{axis}[
    xlabel={copy length $k$},
    ylabel={speed-up},
    xmin=4, xmax=18,
    ymin=1.7, ymax=2.5,
    xtick={4,6,...,18},
    ytick={1.7,1.8,...,2.5},
    grid=major,
    grid style={dashed,gray!30},
    legend pos=south east,
]

\addplot[
    color=blue,
    mark=square,
    ]
    coordinates {
    (4,2.03)(5,2.14)(6,2.22)(7,2.27)(8,2.31)(9,2.34)(10,2.36)(11,2.38)(12,2.41)(13,2.45)(14,2.47)(15,2.46)(16,2.45)(17,2.46)(18,2.46)
    };
    \addlegendentry{n=1}

\addplot[
    color=red,
    mark=triangle,
    ]
    coordinates {
    (4,1.97)(5,2.06)(6,2.13)(7,2.19)(8,2.22)(9,2.25)(10,2.27)(11,2.29)(12,2.32)(13,2.34)(14,2.36)(15,2.35)(16,2.36)(17,2.38)(18,2.36)
    };
    \addlegendentry{n=2}

\addplot[
    color=green,
    mark=diamond,
    ]
    coordinates {
    (4,1.81)(5,1.88)(6,1.94)(7,1.96)(8,2)(9,2.03)(10,2.05)(11,2.07)(12,2.08)(13,2.11)(14,2.12)(15,2.13)(16,2.13)(17,2.15)(18,2.15)
    };
    \addlegendentry{n=3}

\end{axis}
        \end{tikzpicture}
        \caption{13B RAG}
    \end{subfigure}
    \medskip
    \begin{subfigure}{0.45\textwidth}
        \centering
        \begin{tikzpicture}[scale=0.6]
            \begin{axis}[
    xlabel={copy length $k$},
    ylabel={speed-up},
    xmin=4, xmax=18,
    ymin=1.6, ymax=2.4,
    xtick={4,5,6,7,8,9,10,11,12,13,14,15,16,17,18},
    ytick={1.6,1.7,1.8,1.9,2.0,2.1,2.2,2.3,2.4},
    grid=major,
    grid style={dashed,gray!30},
    legend pos=south east,
]

\addplot[
    color=blue,
    mark=square,
    ]
    coordinates {
 (4, 1.94)
(5, 2.03)
(6, 2.1)
(7, 2.16)
(8, 2.19)
(9, 2.21)
(10, 2.23)
(11, 2.26)
(12, 2.27)
(13, 2.27)
(14, 2.29)
(15, 2.3)
(16, 2.26)
(17, 2.27)
(18, 2.27)
    };
    \addlegendentry{n=1}

\addplot[
    color=red,
    mark=triangle,
    ]
    coordinates {
 (4, 1.89)
(5, 1.99)
(6, 2.02)
(7, 2.06)
(8, 2.07)
(9, 2.13)
(10, 2.12)
(11, 2.09)
(12, 2.13)
(13, 2.15)
(14, 2.16)
(15, 2.18)
(16, 2.1)
(17, 2.09)
(18, 2.17)
    };
    \addlegendentry{n=2}

\addplot[
    color=green,
    mark=diamond,
    ]
    coordinates {
    (4, 1.69)
(5, 1.76)
(6, 1.78)
(7, 1.83)
(8, 1.84)
(9, 1.86)
(10, 1.87)
(11, 1.88)
(12, 1.88)
(13, 1.89)
(14, 1.91)
(15, 1.99)
(16, 1.98)
(17, 2)
(18, 2.01)
    };
    \addlegendentry{n=3}

\end{axis}
        \end{tikzpicture}
        \caption{7B CAG}
    \end{subfigure}
    \centering
    \begin{subfigure}{0.45\textwidth}
        \centering
        \begin{tikzpicture}[scale=0.6]
            \begin{axis}[
    xlabel={copy length $k$},
    ylabel={speed-up},
    xmin=4, xmax=18,
    ymin=1.6, ymax=2.4,
    xtick={4,5,6,7,8,9,10,11,12,13,14,15,16,17,18},
    ytick={1.6,1.7,1.8,1.9,2.0,2.1,2.2,2.3,2.4},
    grid=major,
    grid style={dashed,gray!30},
    legend pos=south east,
]

\addplot[
    color=blue,
    mark=square,
    ]
    coordinates {
 (4,2.06)(5,2.14)(6,2.19)(7,2.23)(8,2.17)(9,2.28)(10,2.28)(11,2.3)(12,2.31)(13,2.36)(14,2.36)(15,2.37)(16,2.29)(17,2.34)(18,2.37)
    };
    \addlegendentry{n=1}

\addplot[
    color=red,
    mark=triangle,
    ]
    coordinates {
(4,1.94)(5,2.01)(6,2.05)(7,2.09)(8,2.12)(9,2.16)(10,2.16)(11,2.18)(12,2.17)(13,2.2)(14,2.2)(15,2.24)(16,2.2)(17,2.23)(18,2.26)
    };
    \addlegendentry{n=2}

\addplot[
    color=green,
    mark=diamond,
    ]
    coordinates {
 (4,1.76)(5,1.81)(6,1.84)(7,1.88)(8,1.9)(9,1.92)(10,1.92)(11,1.94)(12,1.95)(13,1.96)(14,1.96)(15,1.96)(16,1.95)(17,2.01)(18,2.04)
    };
    \addlegendentry{n=3}

\end{axis}
        \end{tikzpicture}
        \caption{13B CAG}
    \end{subfigure}
    \caption{Effect of match length ($n$) and copy length ($k$) on the dev set.}
    \label{fig:kn_speedup}
\end{figure}

\subsection{Effect of Match and Copy Length}
We study the effect of match and copy lengths $n$ and $k$ using the dev set.
As can be seen in Figure~\ref{fig:kn_speedup}, aggressive triggering ($n=1$) and longer copy length gives larger speed-up across different settings, with the gains plateaued when copy length $k$ grows past 15.

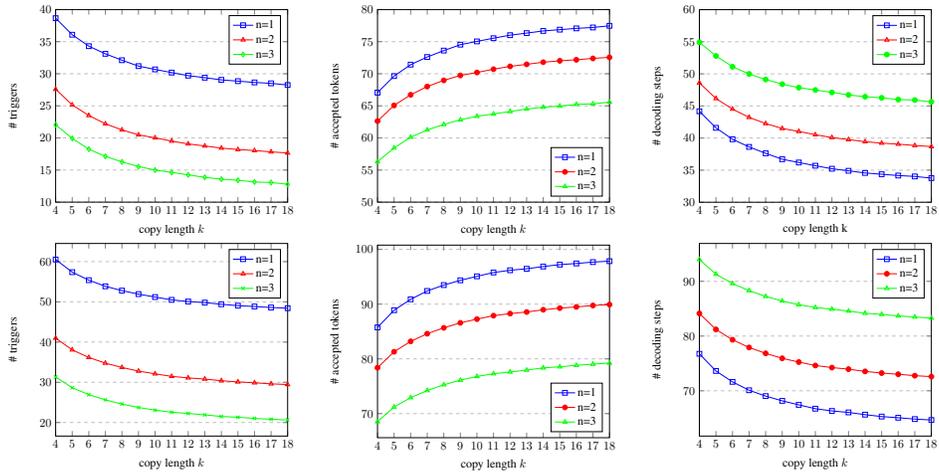
\begin{figure}[htbp]
    \centering
    \begin{subfigure}{0.3\textwidth}
        \centering
        \begin{tikzpicture}[scale=0.45]
            \begin{axis}[    xlabel={copy length $k$},    ylabel={\# triggers},    xmin=4, xmax=18,    ymin=10, ymax=40,    xtick={4,5,...,18},    ytick={10,15,...,40},    legend pos=north east,    ymajorgrids=true,    grid style=dashed,]

\addplot[    color=blue,    mark=square,    ]
    coordinates {
   (4, 38.66)
				(5, 36.08)
				(6, 34.3)
				(7, 33.11)
				(8, 32.11)
				(9, 31.19)
				(10, 30.68)
				(11, 30.18)
				(12, 29.71)
				(13, 29.39)
				(14, 29.05)
				(15, 28.85)
				(16, 28.65)
				(17, 28.51)
				(18, 28.25)
    };
    
\addplot[    color=red,    mark=triangle,    ]
    coordinates {
    (4, 27.59)
				(5, 25.15)
				(6, 23.5)
				(7, 22.21)
				(8, 21.25)
				(9, 20.48)
				(10, 20.02)
				(11, 19.51)
				(12, 19.07)
				(13, 18.75)
				(14, 18.42)
				(15, 18.2)
				(16, 18.04)
				(17, 17.83)
				(18, 17.66)
    };
    
\addplot[    color=green,    mark=diamond,    ]
    coordinates {
    (4, 22.05)
				(5, 19.93)
				(6, 18.25)
				(7, 17.13)
				(8, 16.27)
				(9, 15.54)
				(10, 15.0)
				(11, 14.63)
				(12, 14.25)
				(13, 13.86)
				(14, 13.58)
				(15, 13.41)
				(16, 13.14)
				(17, 13.06)
				(18, 12.78)
    };
    
\legend{n=1,n=2,n=3}
\end{axis}
        \end{tikzpicture}
    \end{subfigure}
    \begin{subfigure}{0.3\textwidth}
        \centering
        \begin{tikzpicture}[scale=0.45]
            \begin{axis}[    xlabel={copy length $k$},    ylabel={\# accepted tokens},    xmin=4, xmax=18,    ymin=50, ymax=80,    xtick={4,5,6,7,8,9,10,11,12,13,14,15,16,17,18},    ytick={50,55,60,65,70,75,80},    legend pos=south east,    ymajorgrids=true,    grid style=dashed,]

\addplot[    color=blue,    mark=square,    ]
    coordinates {
    (4, 67.06)(5, 69.64)(6, 71.42)(7, 72.61)(8, 73.61)(9, 74.53)(10, 75.04)(11, 75.54)(12, 76.01)(13, 76.33)(14, 76.67)(15, 76.87)(16, 77.07)(17, 77.21)(18, 77.47)
    };
    \addlegendentry{n=1}

\addplot[    color=red,    mark=*,    ]
    coordinates {
    (4, 62.63)(5, 65.07)(6, 66.72)(7, 68.01)(8, 68.97)(9, 69.74)(10, 70.2)(11, 70.71)(12, 71.15)(13, 71.47)(14, 71.8)(15, 72.02)(16, 72.18)(17, 72.39)(18, 72.56)
    };
    \addlegendentry{n=2}

\addplot[    color=green,    mark=triangle,    ]
    coordinates {
    (4, 56.32)(5, 58.44)(6, 60.12)(7, 61.24)(8, 62.1)(9, 62.83)(10, 63.37)(11, 63.74)(12, 64.12)(13, 64.51)(14, 64.79)(15, 64.96)(16, 65.23)(17, 65.31)(18, 65.59)
    };
    \addlegendentry{n=3}

\end{axis}
        \end{tikzpicture}
    \end{subfigure}
    \begin{subfigure}{0.3\textwidth}
        \centering
        \begin{tikzpicture}[scale=0.45]
            \begin{axis}[    xlabel={copy length k},    ylabel={\# decoding steps},    xmin=4, xmax=18,    ymin=30, ymax=60,    xtick={4,5,6,7,8,9,10,11,12,13,14,15,16,17,18},    ytick={30,35,40,45,50,55,60},    legend pos=north east,    ymajorgrids=true,    grid style=dashed,]

\addplot[color=blue,mark=square]
    coordinates {
    (4, 44.17)(5, 41.59)(6, 39.81)(7, 38.62)(8, 37.62)(9, 36.7)(10, 36.19)(11, 35.69)(12, 35.22)(13, 34.9)(14, 34.56)(15, 34.36)(16, 34.16)(17, 34.02)(18, 33.76)
    };
\addplot[color=red,mark=triangle]
    coordinates {
    (4, 48.6)(5, 46.16)(6, 44.51)(7, 43.22)(8, 42.26)(9, 41.49)(10, 41.03)(11, 40.52)(12, 40.08)(13, 39.76)(14, 39.43)(15, 39.21)(16, 39.05)(17, 38.84)(18, 38.67)
    };
\addplot[color=green,mark=*]
    coordinates {
    (4, 54.91)(5, 52.79)(6, 51.11)(7, 49.99)(8, 49.13)(9, 48.4)(10, 47.86)(11, 47.49)(12, 47.11)(13, 46.72)(14, 46.44)(15, 46.27)(16, 46.0)(17, 45.92)(18, 45.64)
    };

\legend{n=1,n=2,n=3}
\end{axis}
        \end{tikzpicture}
    \end{subfigure}
    \medskip
    \begin{subfigure}{0.3\textwidth}
        \centering
        \begin{tikzpicture}[scale=0.45]
            \begin{axis}[    xlabel={copy length $k$},    ylabel={\# triggers},    xmin=4, xmax=18,       xtick={4,5,...,18},    legend pos=north east,    ymajorgrids=true,    grid style=dashed,]

\addplot[
    color=blue,
    mark=square,
    ]
    coordinates {
    (4,60.52)(5,57.38)(6,55.39)(7,53.86)(8,52.79)(9,51.92)(10,51.19)(11,50.49)(12,50.08)(13,49.81)(14,49.4)(15,49.05)(16,48.84)(17,48.59)(18,48.42)
    };
\addlegendentry{n=1}

\addplot[
    color=red,
    mark=triangle,
    ]
    coordinates {
    (4,40.98)(5,38.06)(6,36.17)(7,34.76)(8,33.69)(9,32.78)(10,32.1)(11,31.47)(12,31.09)(13,30.8)(14,30.39)(15,30.09)(16,29.87)(17,29.61)(18,29.44)
    };
\addlegendentry{n=2}

\addplot[
    color=green,
    mark=x,
    ]
    coordinates {
    (4,31.32)(5,28.65)(6,26.93)(7,25.64)(8,24.58)(9,23.74)(10,23.07)(11,22.56)(12,22.25)(13,21.9)(14,21.5)(15,21.3)(16,21.02)(17,20.81)(18,20.62)
    };
\addlegendentry{n=3}

\end{axis}
        \end{tikzpicture}
    \end{subfigure}
    \begin{subfigure}{0.3\textwidth}
        \centering
        \begin{tikzpicture}[scale=0.45]
            \begin{axis}[    xlabel={copy length $k$},    ylabel={\# accepted tokens},    xmin=4, xmax=18,    xtick={4,5,6,7,8,9,10,11,12,13,14,15,16,17,18},    legend pos=south east,    ymajorgrids=true,    grid style=dashed,]

\addplot[    color=blue,    mark=square,    ]
    table[    x index=0,    y index=1,    ]{data/cag_accepted_tokens_data.txt};
    \addlegendentry{n=1}
    
\addplot[    color=red,    mark=*,    ]
    table[    x index=0,    y index=2,    ]{data/cag_accepted_tokens_data.txt};
    \addlegendentry{n=2}
    
\addplot[    color=green,    mark=triangle,    ]
    table[    x index=0,    y index=3,    ]{data/cag_accepted_tokens_data.txt};
    \addlegendentry{n=3}

\end{axis}
        \end{tikzpicture}
    \end{subfigure}
    \begin{subfigure}{0.3\textwidth}
        \centering
        \begin{tikzpicture}[scale=0.45]
            \begin{axis}[    xlabel={copy length $k$},    ylabel={\# decoding steps},    xmin=4, xmax=18,    xtick={4,5,6,7,8,9,10,11,12,13,14,15,16,17,18},    legend pos=north east,    ymajorgrids=true,    grid style=dashed,]

\addplot[    color=blue,    mark=square,    ]
    table[    x index=0,    y index=1,    ]{data/cag_decoding_steps_data.txt};
    \addlegendentry{n=1}
    
\addplot[    color=red,    mark=*,    ]
    table[    x index=0,    y index=2,    ]{data/cag_decoding_steps_data.txt};
    \addlegendentry{n=2}
    
\addplot[    color=green,    mark=triangle,    ]
    table[    x index=0,    y index=3,    ]{data/cag_decoding_steps_data.txt};
    \addlegendentry{n=3}

\end{axis}
        \end{tikzpicture}
    \end{subfigure}
    \caption{Decoding statistics with varying match length $n$ and copy length $k$ on the dev set. All statistics are shown on a per sample basis. The first row shows retrieval-augmented scenario and the second row shows cache-assisted scenario. The first column shows the triggering number of copy mechanism; the second column shows the number of copied tokens accepted after verification; the third column shows the decoding steps performed during LLMA decoding. }
    \label{fig:decode_stats}
\end{figure}

To further understand LLMA decoding behaviour, we collect the following decoding statistics on the dev set:
(1) the number of triggering for the copy mechanism per sample;
(2) the number of copied tokens accepted by the verification per sample;
(3) the decoding steps required per sample.
As can be seen in Figure~\ref{fig:decode_stats}, smaller match length $n$ allows more triggering of copying, which leads to more accepted tokens and less decoding steps.
Larger copying length $k$ decreases the number of triggering because consecutive small copying steps for small $k$ might be merged in one large copying step for large $k$.
Larger $k$ also leads to more accepted tokens and less decoding steps.
As noted in Figure~\ref{fig:kn_speedup}, fewer decoding steps by larger $k$ give larger speed-up until $k$ reaches 15 despite larger $k$ wastes more computations when the copied tokens are not accepted. The advantages of fewer decoding steps are 1) better utilization of parallel computing of GPUs and 2) for larger models, fewer data transfers and synchronizations between GPUs. 

\section{Conclusion}\label{sec:conclusion}
In this work, we propose LLMA, a new inference-with-reference decoding method to accelerate LLM generations.
LLMA exploits the overlaps between generation contexts and outputs which naturally occur in several important LLM deployment scenarios such as retrieval-augmented generation, cache-assist generation and multi-turn conversations.
LLMA is easy to deploy and requires no additional models.
Experiments demonstrate the effectiveness of our method, achieving $2\times$ speed-up across different model sizes and application scenarios.

\bibliography{llma}
\bibliographystyle{iclr2022_conference}

\appendix

\section{Prompt Templates}\label{ap_sec:prompt}
The prompt template to get retrieval-augmented generation results from davinci-003 model is in Figure~\ref{fig:dv3_temp}.

\begin{figure}[htbp]
\centering
\begin{quote}
    Respond to the queries according to the presented documents. The query and documents are given in a json string. Here are some guidelines for your response:
    
    1. The response should be informative, visual , logical and actionable.
    
    2. The response should be positive, interesting, entertaining and engaging.
    
    3. The response should avoid being vague, controversial or off-topic.
    
    4. The logics and reasoning in the response should be rigorous, intelligent and defensible.
    
    5. The response can provide additional relevant details to respond thoroughly and comprehensively to cover multiple aspects in depth.
    
    6. The presented documents may be incomplete or irrelevant to the query. The response shouldn't make assumptions on presented documents beyond what's presented.
    
    7. If the presented documents do not contain sufficient information to answer the query completely, the response should use only facts in the presented documents and should not add any information by itself.

    docs: [
        
        \textit{\{doc 1\}}

        ...

        \textit{\{doc n\}}]

    query: \textit{\{query\}}

    response:
\end{quote}
\caption{Prompt template to get generate retrieval-augmented generation from davinci-003.}
\label{fig:dv3_temp}
\end{figure}
A simplified prompt template used in our decoding experiments for our models is in Figure~\ref{fig:simple_temp}.
\begin{figure}[h]
\centering
\begin{quote}

    docs:
        
        \textit{\{doc 1\}}

        ...

        \textit{\{doc n\}}

    query: \textit{\{query\}}

    answer:
\end{quote}
\caption{Prompt template used in decoding experiments for our models.}
\label{fig:simple_temp}
\end{figure}
\end{document}